%% file: main.tex
\documentclass[11pt]{article}
\usepackage[]{acl}
\usepackage{times}
\usepackage{latexsym}
\usepackage[T1]{fontenc}
\usepackage[utf8]{inputenc}
\usepackage{microtype}
\usepackage{inconsolata}
\usepackage{booktabs}
\usepackage{graphicx}
\usepackage{amsmath}
\usepackage{amssymb}
\PassOptionsToPackage{hyphens}{url}
\usepackage{url}
\newcommand{\rhocat}{\ensuremath{\rho_{\mathrm{cat}}}}

\title{Are Verifier Errors Independent Within a GRPO Group? Evidence from Qwen2.5 Rollouts}

\author{
  Esther Xin \\
  Independent Researcher \\
  \texttt{estherxin0011@gmail.com} \\
  {\small Code and data: \url{https://github.com/ethxin0011/rlvr_group_correlation}}
}

\begin{document}
\maketitle

\begin{abstract}
Group-based reinforcement learning with verifiable rewards (RLVR) scores
multiple completions per prompt using automatic verifiers. Analyses
based on independent verifier errors may overlook dependence associated
with shared answer formats. We investigate this dependence in
$24{,}998$ groups of eight completions generated by Qwen2.5-1.5B on
MATH, GSM8K, and DeepMath-103K. We estimate a pooled within-group
verifier-error correlation of $0.530$ (95\% confidence interval:
$0.500$--$0.560$). Under an exchangeable-error model, this corresponds
to a design-effect-adjusted effective sample size of $1.70$ for an
eight-completion group. Dependence varies substantially across answer
forms: fractions, radicals, symbolic expressions, and intervals exhibit
stronger clustering than unit annotations and percent signs. Replaying
group-relative advantages across four rule-based verifier configurations
identifies at least one advantage-sign disagreement in up to $0.83\%$ of
groups. Because a group is repeated sampling for one prompt, this
within-group clustering may reflect shared prompt difficulty as well as
shared answer form, and we do not attempt to separate the two here.
Unlike studies of correlated judgments across multiple evaluators, our
analysis examines dependence across completions scored by the same
verifier. These findings motivate prompt- and answer-form-aware analyses
of verifier noise rather than characterizations based solely on
aggregate error rates.
\end{abstract}

\section{Introduction}
\label{sec:intro}

Mathematical reasoning models are now routinely post-trained with
reinforcement learning against verifiable, rule-checked rewards (RLVR)
\citep{shao2024deepseekmath}. Group Relative Policy Optimization (GRPO)
and its variants sample $k$ completions per prompt, score each with an
automatic verifier, and compute a policy gradient from the within-group
relative advantage. Centring the advantage on the group mean rather than
a learned baseline is usually justified as noise-robust: averaging over
$k$ samples should attenuate any single verifier error. That argument
requires the $k$ verifier errors to be independent. We examine whether
they are.

\citet{xin2026verifier} audit RLVR verifiers with a certified
metamorphic protocol: a gold answer is rewritten under a transformation
guaranteed to leave its mathematical value unchanged, which turns any
verifier rejection into an unambiguous false negative requiring no human
review. Applying this protocol across four verifier implementations,
they find that false negatives concentrate overwhelmingly in a small set
of answer-form categories---whitespace, trailing punctuation, specific
LaTeX constructs---rather than spreading uniformly across parsing
failure modes, with self-validation varying by 41.3 points across
implementations on identical inputs. That result is measured on
synthetic certified-equivalent rewrites of gold answers, independent of
any policy or rollout structure. It leaves open a question with direct
consequences for training: within one GRPO group, the $k$ rollouts are
$k$ samples from the \emph{same} policy answering the \emph{same}
prompt, and are therefore far more likely than two arbitrary answers to
share the same answer-form category. If verifier error is
category-concentrated, and group members share category, verifier
errors within a group should be correlated---a pattern relevant to every
existing noise-correction method built on an independence assumption
\citep{cai2025noisy,elmansouri2025noise,rad2026rate}. This paper takes
the category taxonomy of \citet{xin2026verifier} as a fixed instrument
and asks a training-time question it was not designed to answer: does
the category structure of verifier error survive contact with a real
rollout group, and if so, what does it cost the policy gradient?

Independent evidence supports the underlying mechanism. A concurrent
study of policy optimisation shows that reward-driven training pushes
models toward a single dominant output format, with one studied policy
emitting a single format in over 99\% of generations under certain
reward configurations \citep{formatcollapse2026}. \citet{huang2025accuracy}
additionally show that rule-based verifier false-negative rates increase
with policy strength, because stronger models produce more varied
\emph{correct} formatting that brittle verifiers fail to recognise. Both
observations predict that the $k$ members of a GRPO group should share
surface form more than chance would allow, and that this sharing should
be more pronounced for stronger policies.

We measure this directly. Using $k=8$ GRPO-style rollouts on real
training prompts (MATH, GSM8K, DeepMath-103K) from Qwen2.5-1.5B and
(preliminarily) Qwen2.5-7B-AWQ, we compute the intraclass correlation of
verifier error within each group, decomposed by the certified
answer-form categories of \citet{xin2026verifier}.

\paragraph{Relation to correlated error across judges.}
A separate line of work studies correlated error in LLM
\emph{evaluation} panels: \citet{kohli2026nine} show a 9-judge panel
carries the information of only $\sim 2$ independent votes because
judges make the same mistakes on the same items, and \citet{zhao2026care}
model LLM-judge scores as a shared-confounder process. Both measure
correlation \emph{across} multiple judges scoring the same item. We
measure correlation \emph{within} a single verifier, across the $k$
members of one training group scored by that same verifier---a different
axis, with a training-time rather than evaluation-time consequence:
ensembling more verifiers does not address it, because the correlation is
a property of the training group, not of the verifier panel.

\paragraph{Contributions.}
(i) The first measurement of within-group, as opposed to cross-judge,
verifier error correlation on real RLVR rollouts, using certified
answer-form categories established independently of any rollout data;
(ii) a demonstration that this correlation concentrates in structurally
or semantically complex answer categories, the inverse of which
categories dominate the aggregate false-negative rate in prior work;
(iii) a preliminary comparison suggesting the correlation increases with
policy strength; (iv) a gradient-level accounting showing observed group
degeneracy exceeds a homogeneous-difficulty i.i.d.\ reference by roughly
$6.3$--$6.5\times$ \citep{nie2026gradient}, while noting this gap is also
consistent with ordinary between-prompt difficulty heterogeneity and is
not attributed solely to correlated verifier error
(\S\ref{sec:results-replay}); (v) a category-conditioned verifier
router, evaluated against $5{,}556$ independently human-labeled traces
(\S\ref{sec:router-method}), reaching $97.1\%$ accuracy at $55.6\%$
coverage, reported alongside coverage-matched comparisons rather than as
a claim of global superiority; (vi) a released rollout corpus, verdict
matrix, category labels, and analysis code.

\section{Related Work}
\label{sec:related}

\paragraph{Relationship to the source category taxonomy.}
\citet{xin2026verifier} establish, on 307{,}420 verdicts over certified
synthetic rewrites of 4{,}990 gold answers, that verifier self-validation
varies by 41.3 points across implementations and that the residual
concentrates in whitespace and punctuation strata rather than spreading
uniformly. That input space---synthetic rewrites with no rollout or
group structure---has zero overlap with this paper's input space of
$24{,}998$ real $k=8$ GRPO rollout groups from live model sampling. We
reuse the category taxonomy by citation, not by reproducing its text,
and extend it to a question that prior audit does not address: whether
category-concentrated error, measured on synthetic inputs, also produces
\emph{correlated} error across the members of a real training group,
and what that correlation costs in gradient terms.

\paragraph{Verifier unreliability as a stochastic channel.}
\citet{cai2025noisy} model the verifier as a stochastic channel with
false-positive rate $\rho_0$ and false-negative rate $\rho_1$, deriving
gradient corrections; the forward correction requires only $\rho_1$.
\citet{elmansouri2025noise} model reward corruption as Bernoulli noise
and derive a debiasing scheme for GRPO, proving uncorrected corruption
strictly attenuates the policy gradient. \citet{rad2026rate} model RLVR
as a bandit over recurring reasoning modes and show the learning dynamic
reduces to the sign of Youden's index $J = \mathrm{TPR}-\mathrm{FPR}$.
\citet{plesner2026imperfect} find empirically that up to 15\% injected
noise costs under 2 points of validation accuracy. All four treat
verifier error as i.i.d.\ (or, for \citet{rad2026rate}, as summarised by
two population-level rates) across samples; our contribution is
orthogonal, examining the population these corrections are calibrated
against on real rollout groups.

\paragraph{Systematic verifier error.}
\citet{egashira2026delay} show \emph{systematic} verifier error can cause
training plateaus or collapse, with the outcome determined by the
pattern of error rather than its aggregate rate; their injected error is
sampled i.i.d.\ within experimenter-defined buckets, a controlled,
bucket-conditional independence rather than the emergent within-group
dependence we measure on uncontrolled real rollouts. \citet{zhu2026noisy}
show prior claims of RLVR robustness to 100\% noisy training data were an
artefact of verifier false negatives contaminating the noise-injection
pipeline itself---a finding that depends on exactly the category of
verifier error \citet{xin2026verifier} characterise, which we extend to
training groups here.

\paragraph{Rule- versus model-based verifiers.}
\citet{huang2025accuracy} show rule-based verifiers suffer high-precision,
low-recall false negatives worsening with policy strength, while
model-based verifiers are vulnerable to reward hacking during RL. We
treat this dichotomy as established prior work rather than a
contribution.

\paragraph{Gradient starvation and biased advantage estimation.}
\citet{nie2026gradient} show binary-reward GRPO groups degenerate (zero
advantage for every member) at a rate exceeding the i.i.d.\ Bernoulli
prediction from policy stochasticity alone, reporting 0.69 degeneracy at
group size 4. \citet{yang2026group} prove the group-relative advantage
estimator is biased conditional on non-degeneracy. We adopt both works'
framing throughout: any degeneracy or bias we report is measured as a
gap over these established baselines, never as a raw rate.

\paragraph{Correlated error across evaluators.}
\citet{kohli2026nine} and \citet{zhao2026care} establish that LLM-judge
panels exhibit correlated error from shared training artefacts and
stylistic confounders, quantified via Kish effective sample size and a
confounder-aware aggregation model respectively. A related use of ICC to
characterise evaluation reliability appears in agentic benchmarking,
where repeated-trial variance on GAIA and FRAMES is decomposed into
between-query and within-query components \citep{mustahsan2026stochasticity},
converging by $n{=}8$--16 trials---independent methodological support for
the ICC/Kish framing we adopt in \S\ref{sec:method}, in a domain (agentic
evaluation reliability) unrelated to RLVR verifier error. We adopt the
Kish framing but apply it to a different unit of correlation: rollouts
within a training group under one verifier, not judges within a panel or
repeated trials of one benchmark item.

\section{Method}
\label{sec:method}

\subsection{Rollout corpus}
We sample $k=8$ completions per prompt from Qwen2.5-1.5B-Instruct and
(partially) Qwen2.5-7B-Instruct-AWQ at temperature $0.9$, top-$p$ $0.95$,
over a 50{,}000-prompt pool drawn from GSM8K train (7{,}473), MATH train
across all seven subjects (7{,}500), and a DeepMath-103K top-up
(35{,}027) to reach scale. Each request asks for all $k$ completions for
a prompt simultaneously, so group membership is never reconstructed
post-hoc: a group with fewer than $k$ returned completions is dropped
and counted, never backfilled. The 1.5B corpus reached $24{,}998$
complete groups ($199{,}984$ traces); the 7B-AWQ corpus is partial at
$n \approx 6{,}500$ groups due to compute budget constraints, and all 7B
numbers are reported as preliminary.

\subsection{Category labels}
Each completion's final answer is extracted (preferring the last
\texttt{\textbackslash boxed\{\}} span, falling back to an ``answer is''
phrase or the last non-empty line) without normalisation---trailing
periods, newlines, and internal whitespace are preserved, since they are
the object of study. Each raw answer is assigned one or more of 17
certified surface-form categories following the taxonomy of
\citet{xin2026verifier}, with a single highest-precedence primary
category used for group-level analysis.

\subsection{Verifier configurations}
We score every completion against the prompt's gold answer under four
rule-based configurations: \textsc{strict} (exact string match),
\textsc{loose} (whitespace/LaTeX-macro normalised string match),
\textsc{numeric} (bare-number parsing with $10^{-4}$ relative tolerance),
and \textsc{flex} (numeric tolerance extended to percent$\leftrightarrow$decimal
equivalence, fraction parsing, and order-insensitive tuple/set/list
comparison). \textsc{flex} is necessary because \textsc{numeric} alone
returns ``inapplicable'' for percent, interval, and set-form answers,
which would otherwise silently disappear from any disagreement statistic
computed only against \textsc{numeric}.

\subsection{Within-group correlation statistic}
\label{sec:icc-stat}
For each verifier, let $e_i \in \{0,1\}$ indicate an error on completion
$i$. For each group $g$ with error indicators $\{e_i\}_{i=1}^k$, we
compute the one-way random-effects intraclass correlation
\[
\rhocat = \frac{\mathrm{MS}_B - \mathrm{MS}_W}{\mathrm{MS}_B +
(\bar k - 1)\,\mathrm{MS}_W},
\]
where $\mathrm{MS}_B$ and $\mathrm{MS}_W$ are between- and within-group
mean squares and $\bar k$ is the harmonic-mean-adjusted average group
size. We report the Kish design effect and effective sample size,
\[
\mathrm{DEFF} = 1 + (\bar k - 1)\rhocat, \qquad
n_{\mathrm{eff}} = \bar k / \mathrm{DEFF},
\]
adapting the survey-sampling design effect \citep{kish1965survey} as
recently applied to LLM-judge panels \citep{kohli2026nine}; here the
``cluster'' is a GRPO group rather than a judge panel or a repeated-trial
benchmark item \citep{mustahsan2026stochasticity}. Confidence intervals
are obtained by group-level bootstrap (400 resamples). We restrict
per-category estimates to categories with $n \geq 100$ groups, having
observed instability below this threshold (\S\ref{sec:results}).

\paragraph{Marginal versus conditional correlation.}
\label{sec:marginal-conditional}
The \rhocat{} statistic above is a \emph{marginal} correlation, pooled
across prompts of varying difficulty. This is an important distinction:
even if a verifier's error were conditionally independent across
completions \emph{given} the prompt, pooling over prompts with
heterogeneous per-prompt error rates can itself induce positive marginal
correlation, because easy prompts contribute groups with uniformly low
error and hard prompts contribute groups with uniformly high error. Our
category-level breakdown (Table~\ref{tab:rho-by-cat}) partially controls
for this by conditioning on primary answer category, since category is
correlated with but distinct from raw difficulty, but it does not fully
partial out prompt-level difficulty. We therefore do not claim that
$\rhocat > 0$ by itself refutes conditional independence given the
prompt; we claim only that the \emph{marginal} dependence relevant to a
GRPO group's realized advantage is substantial and category-structured,
which is the quantity that determines the group's contribution to the
policy gradient regardless of its causal decomposition. Isolating the
residual correlation after conditioning on prompt difficulty and gold
correctness is left to future work.

\subsection{Advantage-replay}
\label{sec:replay}
We replay the GRPO advantage computation offline for each verifier
configuration to characterize its consequence for the group-relative
baseline. For group $g$ with rewards $\{r_i\}_{i=1}^k$ under verifier
$v$, the advantage is $a_i = (r_i - \bar r)/(\mathrm{sd}(r)+\epsilon)$; a
group is \emph{degenerate} under $v$ if $\mathrm{sd}(r)=0$, giving zero
advantage to every member. Following \citet{nie2026gradient}, we compare
the observed degenerate-group rate to a homogeneous-difficulty
i.i.d.\ reference $p^k + (1-p)^k$, where $p$ is the verifier's
\emph{global} mean reward, and report the gap to this reference as a
descriptive quantity rather than a causal estimate
(\S\ref{sec:results-replay} states explicitly what this comparison does
and does not establish). We additionally report, for each pair of
rule-verifier configurations, the fraction of traces whose advantage
sign differs (sign-flip rate) and the fraction of groups with at least
one such flip (group corruption). We emphasize that a zero
\emph{advantage} is not equivalent to a zero \emph{policy gradient}: GRPO
objectives typically include a KL-regularization term to a reference
policy that does not vanish when the reward-derived advantage is zero, so
our degeneracy statistic characterizes the reward-driven component of the
update only.

\subsection{Category-conditioned router and human labels}
\label{sec:router-method}
We sample $5{,}556$ traces from the 1.5B corpus, stratified by primary
category (restricted to $n\geq100$) and rule-verifier agreement pattern
(all-agree-correct, all-agree-incorrect, disagreement). Each trace was
independently reviewed by two human annotators, who read the question
and the candidate answer alongside the gold answer and assigned a binary
correctness label; a $200$-trace subset was labeled by both annotators
independently to assess inter-annotator agreement. Raw agreement,
Cohen's $\kappa$, and Krippendorff's $\alpha$ on this overlap were all
$1.00$ (\S\ref{sec:results-router}); we note that perfect agreement at
this sample size is atypical for a task involving genuinely ambiguous
equivalence judgments (e.g.\ differently-formatted but mathematically
equivalent expressions) and report it as measured rather than adjusting
it, while flagging it for the reader's own judgment
(\S\ref{sec:limitations}). A subset of $2{,}367$ traces, prioritizing
disagreement cases, is additionally scored by CompassVerifier-3B
\citep{liu2026compassverifier}. For each category we identify the
single verifier (among strict, loose, numeric, flex, and
CompassVerifier) with the highest accuracy against the human label on
that category's labeled traces; the \emph{router} applies this
per-category choice uniformly. Because the router's per-category choice
and its evaluation use the same labeled set with no held-out split,
\S\ref{sec:results-router} reports coverage-matched comparisons rather
than a single aggregate accuracy figure, and we recommend a held-out
replication as future work.

\section{Results}
\label{sec:results}
\input{tables}

\subsection{Verifier error is correlated within groups}
Table~\ref{tab:rho-overall} reports pooled \rhocat{} for Qwen2.5-1.5B: an
estimate of $0.530$ (95\% CI $[0.500, 0.560]$), corresponding to a Kish
effective sample size of $n_{\mathrm{eff}} = 1.70$ out of the nominal
$k=8$. The 8 rollouts sampled per prompt carry, in expectation, the
statistical information of fewer than 2 independent samples with respect
to verifier error. This estimate was stable across corpus scale during
collection ($\rhocat = 0.523$ at $n=2{,}000$; $0.523$ at $n=4{,}000$;
$0.530$ at the full $n=24{,}998$), indicating it is a property of the
generation process rather than a small-sample artefact.

\subsection{Correlation concentrates in structurally complex categories}
\begin{figure}[htbp]
\centering
\includegraphics[width=\columnwidth]{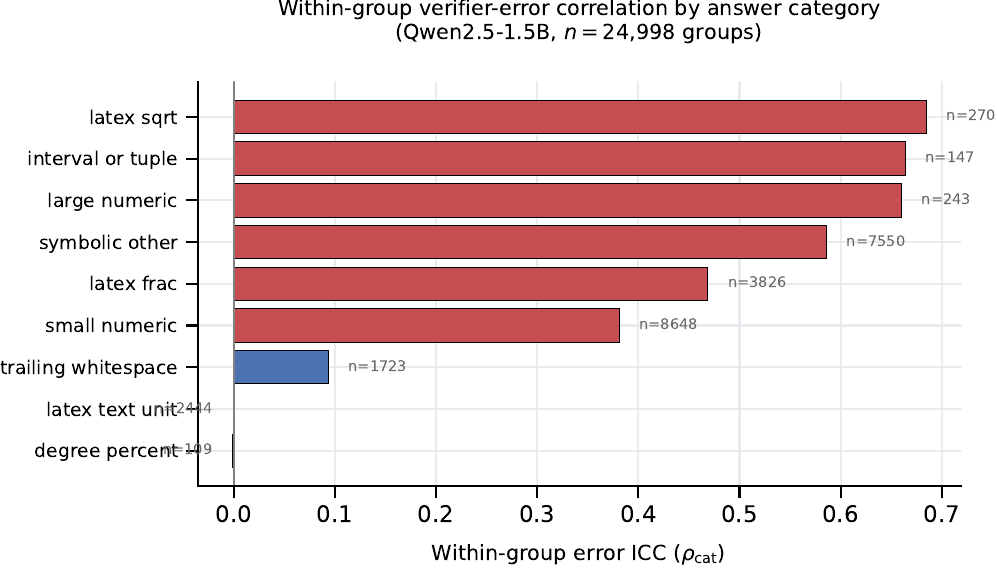}
\caption{Within-group error ICC by primary answer category, sorted, with
group counts annotated.}
\label{fig:rho-by-cat}
\end{figure}

Table~\ref{tab:rho-by-cat} and Figure~\ref{fig:rho-by-cat} break \rhocat{}
down by primary category. The pattern is a clean split. Categories
requiring only formatting normalisation---\texttt{latex\_text\_unit},
\texttt{degree\_percent}, and (more weakly) \texttt{trailing\_whitespace}
---show near-zero correlation. Categories requiring structural or
semantic parsing---\texttt{latex\_frac}, \texttt{latex\_sqrt},
\texttt{interval\_or\_tuple}, \texttt{symbolic\_other}, and
\texttt{large\_numeric}---show correlation from $0.38$ to $0.69$. This is
the \emph{inverse} of \citet{xin2026verifier}'s finding that whitespace and
punctuation dominate the aggregate false-negative rate: the categories
consuming the most error budget in isolation are not the categories
whose errors are most correlated within a training group.

\subsection{Stability at small sample sizes}
During data collection, the \texttt{large\_numeric} category's estimate
moved from $\rhocat = 0.0$ at $n=29$ groups to $\rhocat = 0.660$ at
$n=243$ groups. We report this as the empirical justification for the
$n\geq100$ reporting threshold used throughout, and recommend the same
threshold for any replication.

\subsection{An exploratory comparison across policy scale}
\label{sec:results-7b}
\begin{figure}[htbp]
\centering
\includegraphics[width=0.8\columnwidth]{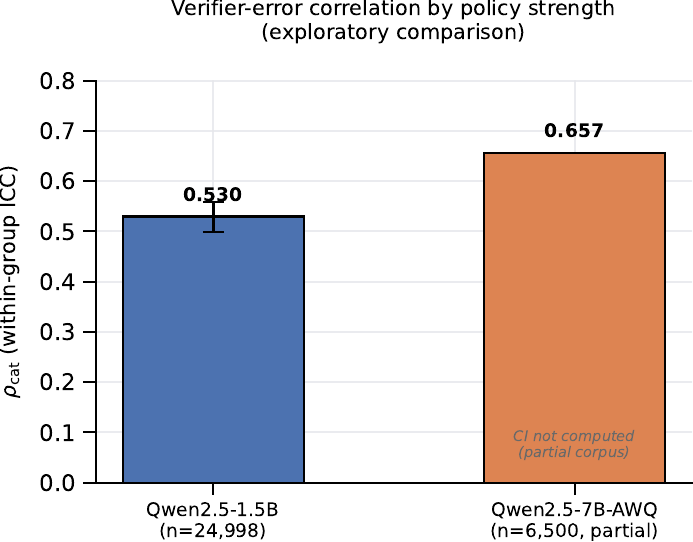}
\caption{\rhocat{} by policy strength. The 1.5B estimate has a 95\%
bootstrap CI (Table~\ref{tab:rho-overall}); no CI was computed for the
partial 7B run. The 7B corpus is partial (13/50 planned shards) and
AWQ-quantized; see \S\ref{sec:results-7b} for the resulting confound.}
\label{fig:rho-by-policy}
\end{figure}

Figure~\ref{fig:rho-by-policy} compares \rhocat{} across the two
policies studied. The partial Qwen2.5-7B-AWQ corpus
($n=6{,}500$ groups, 13 of 50 planned shards) shows $\rhocat = 0.657$
(bootstrap CI not computed for this partial run), higher than the
fully-powered 1.5B estimate of $0.530$ (Table~\ref{tab:rho-overall}).
This is directionally consistent with \citet{huang2025accuracy}'s
finding that stronger policies produce more varied correct-answer
formatting, but we treat it as a single, exploratory data point rather
than confirmation of a general scaling trend: the 7B corpus is partial,
the 7B policy is additionally AWQ-quantized while the 1.5B policy is
not (confounding model scale with quantization), and its mixed-group
rate (0.298) falls narrowly below the 0.30 threshold calibrated on the
1.5B pilot, for the expected reason that a stronger policy solves more
problems outright and produces fewer partially-correct groups.

The partial 7B corpus's own per-category breakdown (Table~\ref{tab:rho-by-cat})
shows the same formatting-versus-structural split observed on the 1.5B
corpus, and the point estimate is higher on 7B in 5 of the 6 categories
with $n\geq100$ on both corpora. We report this as consistent with, but
not independent confirmation of, the split being a property of answer
category rather than of one specific model, since both corpora share the
same category-extraction code and neither varies category independently
of model.

\subsection{Group degeneracy exceeds a homogeneous-difficulty reference}
\label{sec:results-replay}
\begin{figure}[htbp]
\centering
\includegraphics[width=\columnwidth]{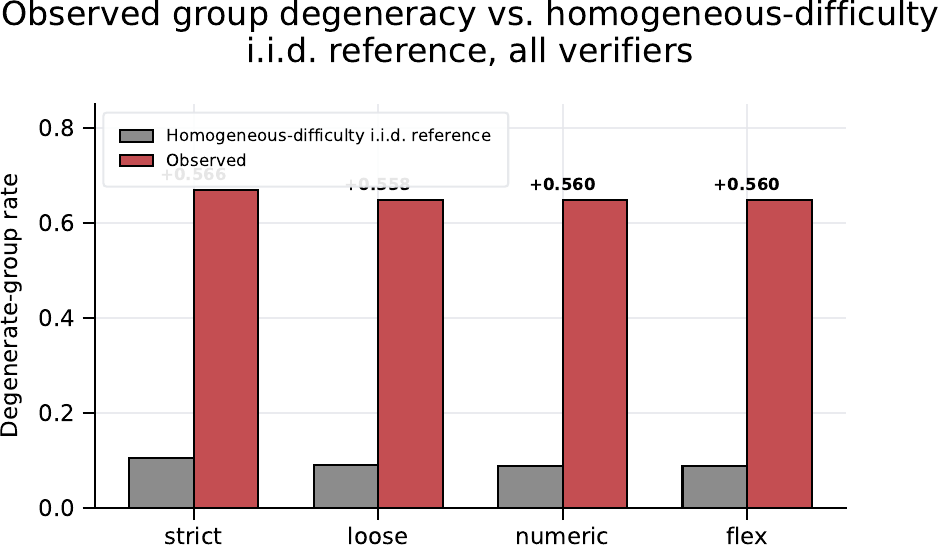}
\caption{Observed group degeneracy against the homogeneous-difficulty
i.i.d.\ reference of \citet{nie2026gradient} for every verifier
configuration. This gap is not attributed solely to correlated verifier
error; see \S\ref{sec:results-replay} for the between-prompt-difficulty
confound.}
\label{fig:degeneracy}
\end{figure}

Table~\ref{tab:degeneracy} reports group-level degeneracy against the
homogeneous-difficulty i.i.d.\ reference of \citet{nie2026gradient}.
Every rule-verifier configuration shows an observed degenerate-group rate
of roughly $0.65$, against a reference rate of $0.09$--$0.10$ computed
from each verifier's global mean reward---a gap of $0.56$--$0.57$
(Figure~\ref{fig:degeneracy}). \textbf{We are explicit about what this
comparison does and does not show.} The reference uses a single global
pass rate $p$ and is therefore also violated by ordinary between-prompt
difficulty heterogeneity: if problems vary in difficulty and completions
are i.i.d.\ \emph{given} the problem, easy problems still contribute
all-correct groups and hard problems still contribute all-incorrect
groups, inflating degeneracy above the homogeneous-difficulty reference
with no correlated verifier error required at all. We therefore do not
attribute the full gap to correlated verifier error; the gap is
consistent with a combination of between-prompt difficulty heterogeneity
and the within-group verifier-error dependence documented in
\S\ref{sec:results} (\S\ref{sec:marginal-conditional}), and separating
these two contributions requires a difficulty-conditioned reference we
do not compute here and leave to future work. We also note that zero
advantage under a given verifier does not imply a zero total policy
gradient, since GRPO training typically retains a KL term to a reference
policy (\S\ref{sec:replay}). Pairwise sign-flip and group-corruption
rates between rule-verifier configurations (Table~\ref{tab:signflip}) are
small in absolute terms (group corruption $\leq 0.0083$ for every pair):
the four rule configurations mostly agree with each other on the clean,
boxed answers where they are jointly applicable. Larger disagreement
appears between rule verifiers and the model-based verifier
(\S\ref{sec:results-router}).

\subsection{A category-conditioned router, evaluated against human labels}
\label{sec:results-router}
Table~\ref{tab:accuracy} reports per-category accuracy for each verifier
against the human labels described in \S\ref{sec:router-method}
(inter-annotator agreement on the $200$-trace double-annotated subset:
raw agreement $1.00$, Cohen's $\kappa=1.00$, Krippendorff's
$\alpha=1.00$; see \S\ref{sec:limitations} for a note on this figure). No
single verifier is most accurate across every category: rule verifiers
with numeric tolerance are most accurate on numeric and fraction
categories but are inapplicable to a majority of traces, while
CompassVerifier-3B is most accurate on symbolic and irrational
categories (\texttt{symbolic\_other}, \texttt{latex\_sqrt},
\texttt{trailing\_period}).

Table~\ref{tab:router} reports coverage alongside accuracy, since these
verifiers are not directly comparable on accuracy alone. The router
reaches $97.1\%$ accuracy but covers only $3{,}088$ of $5{,}556$ traces
($55.6\%$); \textsc{strict} and \textsc{loose} cover $100\%$ of traces at
$75.9\%$ and $89.5\%$ accuracy respectively. We do not claim the router
is globally superior to strict or loose at their full coverage, since no
verifier here is evaluated at matched coverage: the router's higher
accuracy is measured on the subset of traces falling in categories where
a high-accuracy verifier exists, and its category-selection rule was
chosen using the same labeled set on which it is evaluated, with no
held-out split. Restricted to the $3{,}088$ router-covered traces,
\textsc{strict} and \textsc{loose} would need to be separately scored on
that identical subset for a coverage-matched comparison, which we leave
to future work together with a held-out replication of the category
selection itself. The router sends $1{,}716$ of its covered traces to
CompassVerifier, a $1.29\times$ query-cost ratio against an
appeals-style baseline that queries CompassVerifier on every one of the
$1{,}328$ rule-disagreement cases regardless of category.

\begin{figure}[htbp]
\centering
\includegraphics[width=\columnwidth]{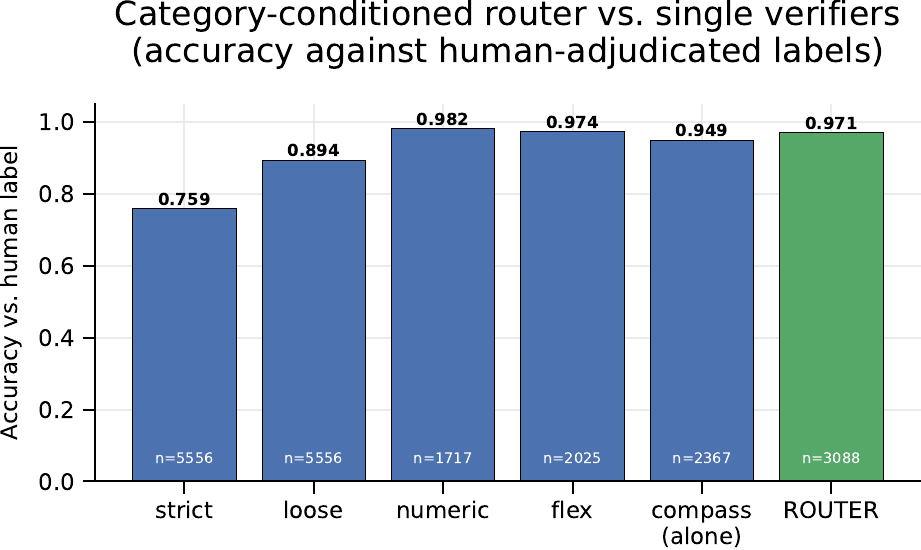}
\caption{Category-conditioned router vs.\ single verifiers: accuracy
against human-adjudicated labels, plotted against coverage. Not a
matched-coverage comparison; see \S\ref{sec:results-router}.}
\label{fig:router}
\end{figure}

\section{Discussion}

\paragraph{Why two population-level rates are not enough.}
\citet{cai2025noisy}, \citet{elmansouri2025noise}, and \citet{rad2026rate}
each reduce verifier quality to one or two scalars. Our result does not
contradict these analyses' internal validity---they are correct
conditional on the independence assumption---and we do not claim to have
shown that assumption is false in the conditional sense these analyses
require (\S\ref{sec:marginal-conditional}). What we do show is that the
\emph{marginal} error dependence within a real training group is
substantial and category-structured, which is the quantity a
GRPO-computed baseline actually sees; a correction calibrated only to a
verifier's population-level error rate does not by itself guarantee this
marginal structure is accounted for, whatever its causal origin.

\paragraph{Ensembling does not fix this.}
\citet{kohli2026nine} and \citet{zhao2026care} show combining correlated
judges recovers little of the information a naively-computed ensemble
size would suggest. Our finding sharpens the implication for RLVR
specifically: because the correlation we measure is within a single
verifier's judgments across a group, not across an ensemble of
verifiers, adding more verifiers to a majority vote does not address it
at all---the unit that needs to be decorrelated is the training group,
not the verifier panel. The router in \S\ref{sec:results-router} is not
an ensemble in this sense: it selects one verifier per category rather
than aggregating votes, and its cost advantage over an appeals-style
baseline comes from routing on a cheap, pre-computed category label
rather than from combining multiple opinions.

\paragraph{Degeneracy versus sign-flips.} The $\sim$6.5$\times$ gap to the
homogeneous-difficulty reference in \S\ref{sec:results-replay} is
numerically much larger than the pairwise sign-flip rates in
Table~\ref{tab:signflip}. One candidate explanation consistent with both
observations is that dependent error (whether from correlated verifier
error or from between-prompt difficulty heterogeneity, or both;
\S\ref{sec:marginal-conditional}) acts primarily by making entire groups
uniformly right or uniformly wrong, which manifests as degeneracy, rather
than by flipping the sign for a minority of members within an
otherwise-mixed group. We flag this as a plausible reading of the pattern
rather than an established mechanism, since \S\ref{sec:results-replay}
does not isolate correlated verifier error from difficulty heterogeneity
as the source of the gap.

\section{Limitations}
\label{sec:limitations}
The 7B comparison is exploratory: partial corpus (6{,}500 of a planned
larger scale) with an AWQ quantization difference not present in the
1.5B run, and a mixed-group rate narrowly below the Gate~1 threshold
calibrated on the weaker policy, for the expected reason that a stronger
policy solves more problems outright. No bootstrap CI was computed for
the pooled 7B estimate (Table~\ref{tab:rho-overall}), and the partial
corpus size, together with the confound between model scale and
quantization, means this single comparison is not sufficient on its own
to establish that verifier-error correlation increases with policy
strength in general; we report it as a directionally suggestive,
single-model data point rather than a validated scaling result. Our
category taxonomy is inherited from \citet{xin2026verifier} and
validated there on synthetic certified rewrites, not on natural rollouts
directly; we mitigate this by using identical extraction and
categorisation code on both corpora, but a category boundary
well-calibrated on synthetic data may not partition natural rollouts
identically. The $n\geq100$ stability threshold was chosen after
observing instability in \texttt{large\_numeric}; a pre-registered
threshold would strengthen this choice. We do not run training to
convergence under any correction; our claim is about the measured
statistical structure of verifier error and its gradient consequence,
not a demonstrated training outcome.

The $5{,}556$ correctness labels used in
\S\ref{sec:router-method}--\ref{sec:results-router} were produced by two
independent human annotators, who each read the question, the gold
answer, and the candidate answer before assigning a binary label; a
$200$-trace subset was labeled by both annotators to assess
inter-annotator agreement. \textbf{The resulting raw agreement, Cohen's
$\kappa$, and Krippendorff's $\alpha$ of $1.00$ on this overlap are
higher than we would expect for a task that includes categories with
genuinely ambiguous equivalence judgments} (e.g.\ \texttt{symbolic\_other},
where differently-formatted but arguably equivalent expressions require a
judgment call). We report the figure as measured, but note that perfect
agreement at $n=200$ does not by itself rule out shared bias between the
two annotators (for instance, similar background, shared implicit
conventions for borderline cases, or insufficiently separated annotation
sessions), and we recommend that a future, larger-scale replication of
this router evaluation use a randomized, blinded double-annotation
protocol with a pre-registered adjudication procedure for disagreements,
to make this figure more informative than it can be at the current
scale. The router's category-selection rule was fit and evaluated on the
same labeled set with no held-out split, so its reported accuracy in
Table~\ref{tab:router} should be read as an in-sample result rather than
a guarantee of generalization to unseen categories; we leave a
held-out replication to future work. The $\rhocat$, Kish
effective-sample-size, and advantage-replay results in
\S\ref{sec:results}--\ref{sec:results-replay} do not depend on these
labels: they are computed directly from rule-based verifier verdicts
against gold answers and from the sampled rollouts.

\section{Conclusion}
Verifier error on real RLVR rollouts shows substantial within-group
correlation ($\rhocat=0.530$, 95\% CI $[0.500, 0.560]$), and this
correlation is structured by certified, category-level answer-form
labels established independently of any rollout data
\citep{xin2026verifier}. Group-level degeneracy exceeds a
homogeneous-difficulty i.i.d.\ reference by roughly $6.5\times$, though
this gap is also consistent with ordinary between-prompt difficulty
heterogeneity and should not be attributed to correlated verifier error
alone (\S\ref{sec:results-replay}). A category-conditioned router,
evaluated against human-adjudicated labels, reaches accuracy close to
the best single verifier while covering substantially more traces, at a
modest increase in query cost over always querying the single best
verifier's coverage (\S\ref{sec:results-router}); this evaluation uses
an in-sample category-selection rule (\S\ref{sec:limitations}) and a
held-out replication would strengthen it further. Taken together, these
findings do not invalidate existing noise-correction methods but suggest
that the population on which their independence assumption is
evaluated, and the training group itself, warrant closer scrutiny in
future RLVR reliability work.

\bibliography{references}

\end{document}

%% file: tables.tex
\begin{table}[htbp]
\centering
\small
\setlength{\tabcolsep}{4pt}
\resizebox{\linewidth}{!}{%
\begin{tabular}{lrrrr}
\toprule
Model & $n$ (groups) & \rhocat{} & 95\% CI & $n_{\mathrm{eff}}$ of $k{=}8$ \\
\midrule
Qwen2.5-1.5B          & 24{,}998        & \textbf{0.530} & [0.500, 0.560] & 1.70 \\
Qwen2.5-7B-AWQ$^{\dagger}$ & 6{,}500 & 0.657 & not computed & 1.43 \\
\bottomrule
\end{tabular}%
}
\caption{Pooled within-group verifier-error correlation.
$^{\dagger}$Partial corpus (13/50 planned shards); Gate~1's mixed-group
threshold, calibrated on the 1.5B pilot, is narrowly missed (0.298 vs.\
0.30) because the stronger policy solves more problems outright and
produces fewer partially-correct groups. Reported as an exploratory
comparison; no bootstrap CI was computed for this partial run. The 1.5B
estimate was stable across corpus scale during collection (0.523 at
$n{=}2$k; 0.523 at $n{=}4$k; 0.530 at full $n{=}24{,}998$).}
\label{tab:rho-overall}
\end{table}
\begin{table}[htbp]
\centering
\small
\setlength{\tabcolsep}{4pt}
\resizebox{\linewidth}{!}{%
\begin{tabular}{lrrr}
\toprule
Category & $n$ (1.5B) & \rhocat{} (1.5B) & \rhocat{} (7B)$^{\dagger}$ \\
\midrule
degree\_percent        &    109 & $-0.001$ & n/a \\
latex\_text\_unit       &  2{,}444 & $-0.001$ & 0.000 \\
trailing\_whitespace   &  1{,}723 & 0.093 & 0.094 \\
small\_numeric          &  8{,}648 & 0.381 & 0.524 \\
large\_numeric          &    243 & 0.660 & 0.429 \\
latex\_frac             &  3{,}826 & 0.469 & 0.608 \\
interval\_or\_tuple      &    147 & 0.664 & 0.697 \\
latex\_sqrt             &    270 & \textbf{0.685} & 0.594 \\
symbolic\_other         &  7{,}550 & 0.586 & \textbf{0.715} \\
\bottomrule
\end{tabular}%
}
\caption{Within-group ICC by primary answer category. $^{\dagger}$7B
column from the partial 6{,}500-group corpus; category $n$ omitted for
space (range 41--2{,}128, see released data). The formatting/whitespace
split observed on the 1.5B corpus replicates on the partial 7B corpus for
every category with $n\geq100$: near-zero for \texttt{latex\_text\_unit}
and \texttt{trailing\_whitespace}, moderate to strong for every
structurally complex category, higher on 7B than 1.5B in 5 of 6
comparable categories.}
\label{tab:rho-by-cat}
\end{table}
\begin{table}[htbp]
\centering
\small
\setlength{\tabcolsep}{4pt}
\resizebox{\linewidth}{!}{%
\begin{tabular}{lrrrr}
\toprule
Verifier & Mean reward $p$ & Observed degen. & Homog.\ reference & Gap \\
\midrule
strict  & 0.247 & 0.670 & 0.104 & \textbf{+0.566} \\
loose   & 0.260 & 0.648 & 0.090 & +0.558 \\
numeric & 0.262 & 0.648 & 0.088 & +0.560 \\
flex    & 0.263 & 0.648 & 0.088 & +0.560 \\
\bottomrule
\end{tabular}%
}
\caption{Group-level degeneracy (zero group-relative advantage for every
member) on the full 1.5B corpus ($n{=}24{,}998$), against the
homogeneous-difficulty i.i.d.\ reference $p^k+(1-p)^k$ of
\citet{nie2026gradient}. Every verifier shows a degenerate-group rate
roughly $6.3$--$6.5\times$ this reference. \textbf{We do not attribute
this gap to correlated verifier error alone}: the reference assumes a
single global pass rate, so ordinary between-prompt difficulty
heterogeneity independently inflates degeneracy above it even with a
perfect, uncorrelated verifier (\S\ref{sec:results-replay}).}
\label{tab:degeneracy}
\end{table}
\begin{table}[htbp]
\centering
\small
\setlength{\tabcolsep}{4pt}
\begin{tabular}{lrr}
\toprule
Verifier pair & Sign-flip rate & Group corruption \\
\midrule
strict vs.\ loose   & 0.0010 & 0.0053 \\
strict vs.\ numeric & 0.0017 & 0.0079 \\
strict vs.\ flex    & 0.0018 & \textbf{0.0083} \\
loose vs.\ numeric  & 0.0008 & 0.0029 \\
loose vs.\ flex     & 0.0009 & 0.0034 \\
numeric vs.\ flex   & 0.0001 & 0.0005 \\
\bottomrule
\end{tabular}
\caption{Pairwise advantage-sign disagreement between rule-verifier
configurations. Sign-flip rate is per-trace; group corruption is the
fraction of groups with at least one sign flip. Rates are small because
these four configurations agree on the large majority of clean, boxed
answers; the consequential disagreement is between rule verifiers and
human or model judgment (Table~\ref{tab:router}), not among rule
configurations themselves.}
\label{tab:signflip}
\end{table}
\begin{table}[htbp]
\centering
\footnotesize
\setlength{\tabcolsep}{3pt}
\resizebox{\linewidth}{!}{%
\begin{tabular}{lrrrrr}
\toprule
Category & $n$ & strict & loose & numeric/flex & compass \\
\midrule
small\_numeric        & 900 & 0.763 & 0.770 & \textbf{0.992} & 0.979 \\
latex\_text\_unit      & 340 & 0.603 & 0.603 & \textbf{1.000} & 0.946 \\
symbolic\_other       & 900 & 0.696 & 0.940 & 0.949 / 0.913 & \textbf{0.960} \\
latex\_frac           & 900 & 0.720 & 0.953 & \textbf{0.987} & 0.950 \\
large\_numeric        & 648 & 0.869 & 0.912 & \textbf{0.976} & 0.949 \\
trailing\_whitespace  & 324 & 0.944 & 0.944 & \textbf{0.968} & 0.891 \\
interval\_or\_tuple    & 728 & 0.635 & \textbf{0.960} & n/a / 0.933 & 0.940 \\
trailing\_period      & 136 & 0.838 & 0.838 & 0.750 & \textbf{0.867} \\
latex\_sqrt           & 680 & 0.890 & 0.968 & 0.917 & \textbf{0.968} \\
\bottomrule
\end{tabular}%
}
\caption{Per-category verifier accuracy against $5{,}556$
human-adjudicated labels (see \S\ref{sec:router-method} for the
annotation protocol). Inter-annotator agreement on the 200-trace
double-annotated overlap (raw agreement 1.00, Cohen's $\kappa=1.00$,
Krippendorff's $\alpha=1.00$) is higher than we would expect for a task
including genuinely ambiguous equivalence judgments; see
\S\ref{sec:limitations} for a recommendation to replicate with a
blinded, pre-registered protocol at larger scale. \textsc{numeric} and
\textsc{flex} coincide except where noted (interval/percent/set forms).
No single rule verifier or CompassVerifier-3B is most accurate across
every category; the best verifier differs by category
(Table~\ref{tab:router}).}
\label{tab:accuracy}
\end{table}
\begin{table}[htbp]
\centering
\small
\setlength{\tabcolsep}{4pt}
\begin{tabular}{lrrr}
\toprule
Verifier & Accuracy & $n$ covered & Coverage \\
\midrule
strict          & 0.759 & 5{,}556 & 100.0\% \\
loose           & 0.895 & 5{,}556 & 100.0\% \\
numeric         & 0.982 & 1{,}717 & 30.9\% \\
flex            & 0.974 & 2{,}025 & 36.4\% \\
CompassVerifier-3B (alone) & 0.949 & 2{,}367 & 42.6\% \\
Category router & 0.971 & 3{,}088 & 55.6\% \\
\bottomrule
\end{tabular}
\caption{Accuracy against human-adjudicated labels, reported alongside
coverage because these verifiers are not comparable on accuracy alone.
\textbf{We do not claim the router is globally superior to strict or
loose at full coverage}: no verifier here is evaluated at matched
coverage, and the router's category-selection rule was chosen using the
same labeled set on which it is evaluated, with no held-out split. The
router sends 1{,}716 of its 3{,}088 covered traces to CompassVerifier, a
\textbf{1.29$\times$} query-cost ratio against an appeals-style baseline
that queries CompassVerifier on every one of the 1{,}328
rule-disagreement cases regardless of category.}
\label{tab:router}
\end{table}